\documentclass{article}

\usepackage{arxiv}

\usepackage[utf8]{inputenc} % allow utf-8 input
\usepackage[T1]{fontenc}    % use 8-bit T1 fonts
\usepackage{hyperref}       % hyperlinks
\usepackage{url}            % simple URL typesetting
\usepackage{booktabs}       % professional-quality tables
\usepackage{amsfonts}       % blackboard math symbols
\usepackage{nicefrac}       % compact symbols for 1/2, etc.
\usepackage{microtype}      % microtypography
\usepackage{lipsum}
\usepackage{graphicx}
\usepackage{bbding} %首先在导言区调用bbding包
\graphicspath{ {./images/} }
\usepackage{color}

\title{Correspondence Learning for Controllable Person Image Generation}

\author{
 Shilong Shen\\
  Harbin Institute of Technology,Shenzhen\\
  \texttt{19S153199@stu.hit.edu.cn} \\
}

\begin{document}
\maketitle
\begin{abstract}
     We present a generative model for controllable person image synthesis,as shown in Figure \ref{two-task-explanation} , which can be applied to pose-guided person image synthesis, $i.e.$, converting the pose of a source person image to the target pose while preserving the texture of that source person image, and clothing-guided person image synthesis, $i.e.$, changing the clothing texture of a source person image to the desired clothing texture. By explicitly establishing the dense correspondence between the target pose and the source image, we can effectively address the misalignment introduced 
     by pose tranfer and generate high-quality images.
     Specifically, we first generate the target semantic map under the guidence of the target pose, which can provide more accurate pose representation and structural constraints during the generation process.     
     Then, decomposed attribute encoder is used to extract the component features, which not only helps to establish a more accurate dense correspondence, but also realizes the clothing-guided person generation. 
     After that, we will establish a dense correspondence between the target pose and the source image within the sharded domain. The source image feature is warped according to the dense correspondence to  flexibly account for deformations. 
     Finally, the network renders image based on the warped source image feature and the target pose.
     Experimental results show that our method is superior to state-of-the-art methods in pose-guided person generation and its effectiveness in clothing-guided person generation.
 
\end{abstract}

% keywords can be removed

\section{Introduction}
Person image generation is regarded as one of the most difficult problems in image analysis, and has important applications in movie making, virtual reality, and data enhancement. Pose-guide person image generation and clothing-guide person image generation are important tasks in this topic, but these tasks are extremely challenging due to the non-rigid nature of the human body and the intricate relationship among various attributes of the human body.

For pose-guided person image generation, proposed by \cite{ma2017pose}, its goal is to convert the pose to the target pose while preserving the texture of the source image. This task is challenging due to the deformation between the source image and the target image, and the occlusion caused by the pose change. The current methods can be roughly divided into two categories: direct methods and warping-based methods. The direct method\cite{ma2017pose,men2020controllable,ma2018disentangled,esser2018variational} uses the target pose, the source pose and the source image as the input of a vanilla convolutional network to directly generate an image. However, images generated in this way often lead to blurred results because the deformation between the source image and the target image is not considered. The warping-based method usually uses affine transformation\cite{siarohin2018deformable}, optical flow\cite{ren2020deep,li2019dense} or attention mechanism\cite{zhu2019progressive} to establish the correspondence between the source image and the target image, and explicitly solves the deformation problem caused by the pose change. Compared with the direct method, it can generate More realistic images. But most of these methods only focus on pose-guided person image generation and cannot perform clothing-guided person image generation, due to the complex relationship between person attributes. 
Different from the appeal method, a few approcaches combines the two tasks by extracting pose representation and texture features through independent encode modules. However, these approcaches cannot accurately render the texture details of the source image under the target view because they fail to model model the intricate interplay of the source image and target image.

To solve the above problems, in this paper, we combine pose-guided person image generation task and clothing-guided person image generation task. Our starting point is to first extract the target pose feature and texture attributes independently, and then establish the dense corresponding at deep feature space , which can effectively solve the misalignment caused by the pose transfer. Our network consists of the following parts: 1) We first use the sub-network guided by the target pose to generate the target semantic map, which can better guide the image generation by providing stronger structural constraints. 2) We use existing semantic segmentation methods to separate the attributes of the source image and map them to the deep feature space, and finally combine these feature codes to build a complete texture code. Then using an independent pose encoder to obtain the target pose code 3) Using the extracted pose code and texture code, dense correspondence can be established. Under the guidence of dense correspondence,we can obtain the warped source image feature 4) Using the lightweight decoder to render the warped source image feature according to the target pose to obtain the final result. In fact, the above network parts  facilitate each other. First of all, decoupling person attribute extraction can not only better extract features to establish more accurate dense correspondences, but also make clothing-guided person generation possible. At the same time, the establishment of the dense corresponding can solve the deformation problem caused by the pose change and generate more realistic images, which is not only beneficial to the pose-guided person image generation task but also beneficial to the clothing-guided person image generation.
Our contributions are mainly as follows: 1) We combine the pose-guided person image generation task with the clothing-guided person image generation task, and learn the dense correspondence between the target pose and the source image through a joint learning method to slove  deep feature maps misalignment. 2) The experiment results show that our method is superior to state-of-the-art methods on the pose-guided person image generation task due to the preservation of the correct structure and accurate texture details, and further verifies that our method is effective on clothing-guided person image generation tasks. 
  
 \begin{figure}[!ht]
 	\centering%设置内容居中
 	\includegraphics[scale=0.6]{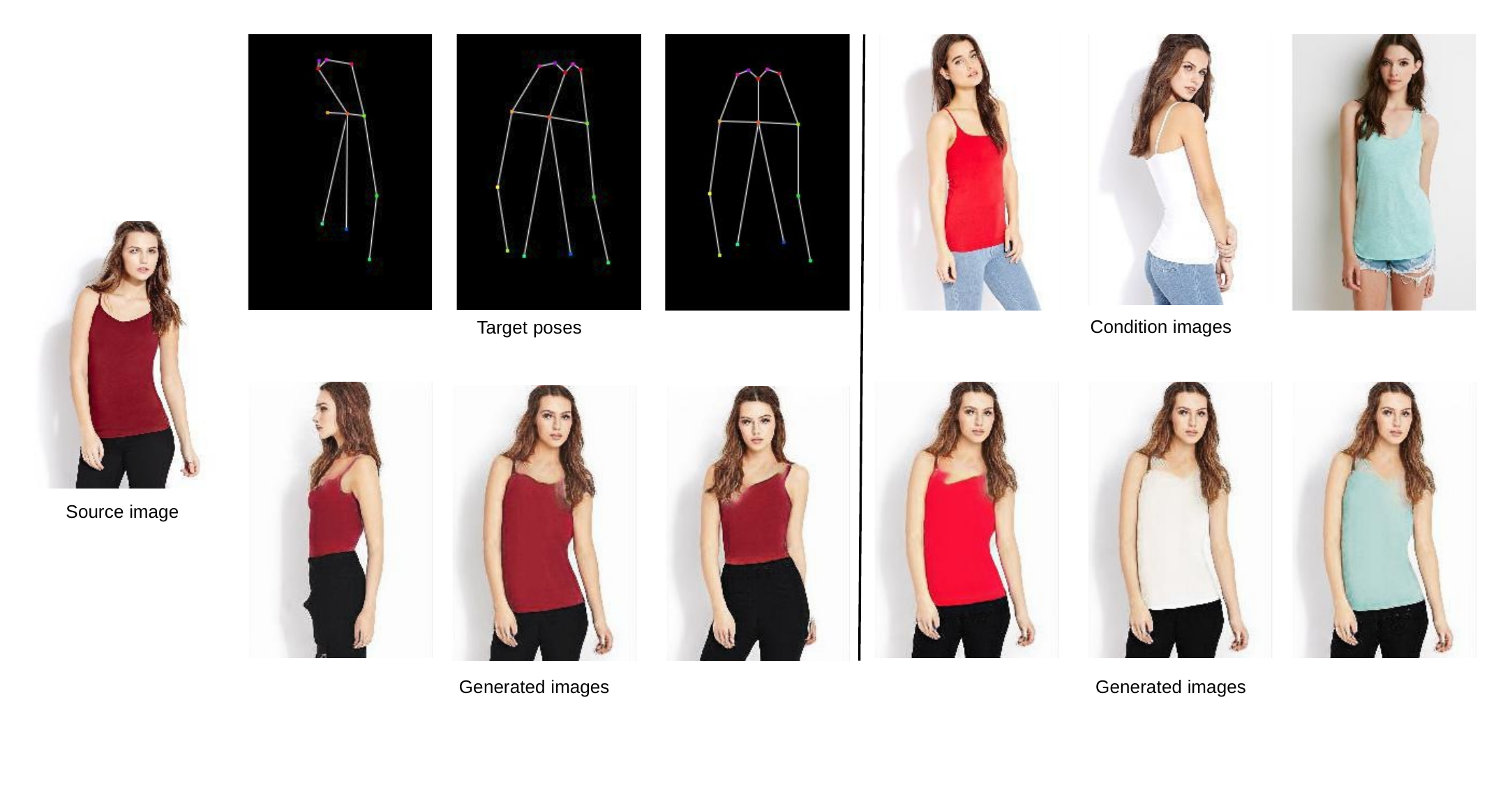}%设置图片的大小和图片的位置
 	\caption{Our method can generate person images in different target poses (left) and transfer upper clothing textures to a person image(right)}%设置图片的名称
 	\label{two-task-explanation}%设置图片的简称
 \end{figure}

\section{ Related Work}
\label{sec:headings}

%\paragraph{Person Image Synthesis}Recent years have witnessed a great breakthrough in image generation by using Generative Adversarial Networks(GAN), Variational Autoencoder(VAE) and other deep generative models. 

%\paragraph{Semantic correspondence}
%\subsection{Person Image Synthesis}
Recent years have witnessed a huge breakthrough in image generation through the use of deep generative models such as Generative Adversarial Networks (GAN) \cite{goodfellow2014generative} and Variational Autoencoder (VAE) \cite{kingma2013auto}. Among them, GAN, including generator and discriminator, has been widely used because of its ability to generate realistic images in adversarial training methods. DCGAN \cite{Radford2016UnsupervisedRL} combines convolutional neural network and GAN to generate realistic images throght a unsupervised method. In practical applications, we often need to generate conditional images, the proposal of CGAN\cite{mirza2014conditional} solves this problem very well, which  allows GAN to generated images conditionally. Pix2pixHD \cite{wang2018high} successfully applies GAN to image translation tasks, which can generate high resolusion images from semantic label maps by multi scale generator and discriminator. Park $et \ al.$ \cite{park2019semantic} proposes a novel network structure called SPADE for image translation task, which effectively avoids "wash away" semantic information. StyleGAN \cite{karras2019style}  successfully applied GAN to the task of face generation through a new generator architecture composed of AdaIN layers, and achieved amazing results. StyleGAN2 \cite{karras2020analyzing} solves the problem of artifacts in the generated image by modifying the AdaIN structure, and further improves the quality of the generated image. However, these methods cannot be effectively applied to the task of person image generation due to the misalignment caused by pose deformation and complexity of appearance. Different from these methods, we establish dense correspondences in the deep feature space to solve the misalignment of deep features and effectively render image textures for person image generation.

Person image generation is an important sub-area of image generation. Many methods have been proposed for pose-guided person image generation.
PG2 \cite{ma2017pose} proposes to use a two-stage network "from coarse to fine" to generate person under the guidance of the source image and the keypoint-based target pose.
Def-GAN \cite{siarohin2018deformable} introduced a U-net with a deformable skip connection to alleviate the problem of deformation.
Li $et \ al.$ \cite{li2019dense} propose to generate a dense 3D appearance flow, which can guide the transfer of pixels between poses.
PATN \cite{zhu2019progressive} introduces the attention mechanism into the generator to solve the problem of deformation in a progressive generation method.
Ren $et \ al.$ \cite{ren2020deep} regards the target image as the deformation version of the source image, and uses a global flow field estimator to calculate correspondence between sources and targets and a local attention sampling module as a content-aware sampling method to generate the image.
There are also some methods that try to combine pose-guided person image generation with clothing-guided person image generation.
Based on VAE, VUnet \cite{esser2018variational} proposed the conditional U-net to  manipulate pose and appearance.
Zhou $et \ al.$ \cite{zhou2019text}uses a text-guided approach to transform the appearance of person,including pose and attribute.
Men $et \ al.$ \cite{men2020controllable} proposed to decouple person attributes by the existing human parsing method, which not only realizes pose transfer but also realizes the control of person attributes.
However, all methods above usually either only focus on pose-guided person  image generation or simply combine pose-guided person image generation and clothing-guided person image generation. Our method solves the problem of deep feature misalignment by establishing the dense correspondence between the target pose and the source image in the deep feature space. A joint learning method that effectively combines pose-guided person image generation task and clothing-guided person image generation to generate high-quality person images.

\section{Method Description}
Our goal is to achieve controllable person image synthesis. Different from previous pose-guided person image generation methods, we also need to consider clothing-guided person image generation. In order to complete this difficult task, we divide the networks architectures into three parts: pose-guided semantic map generation, generator and discriminator, due to the powerful ability of GAN in image generation. firstly, we present pose-guided semantic map generation. Then, generator and discriminator will be discussed in details, respectively. Finally, we will describle in detail the objective function used in our network.

we give some notations here. $\{(I_s^{(i)} \in \mathbb{R}^{3 \times H \times W},I_t^{(i)}\in \mathbb{R}^{3 \times H \times W})\}_{i=1,...,N}$  denotes pairs of source-target images of same person in different poses. $H$ is the height of the image, $W$ is the width of the image, $N$ is the number of person images. For each pair $(I_s,I_t)$, a keypoint-based source and a keypoint-based target pose $P_s\in \mathbb{R}^{18 \times H \times W}$ and $P_t\in \mathbb{R}^{18 \times H \times W}$ is extracted from the correspondence image, which 
encoder pose as 18-channel heatmap representing 18 joints of a human body. We adopt the same Pose Estimator \cite{cao2017realtime} used in \cite{ma2017pose}. What's more, we extracte semantic map pairs $(S_s,S_t)$ of image pairs $(I_s,I_t)$ with an state-of-the-art Human parser \cite{gong2017look} and converts semantic map $S_s$ into $K$-channel heatmap $M_s \in \mathbb{R}^{K \times H \times W}$ as \cite{men2020controllable}, each channel is a mask map, corresponding to an attribute of the human body.  

\subsection{pose-guided semantic map generation}
\begin{figure}[!ht]
	\centering%设置内容居中
	\includegraphics[scale=0.7]{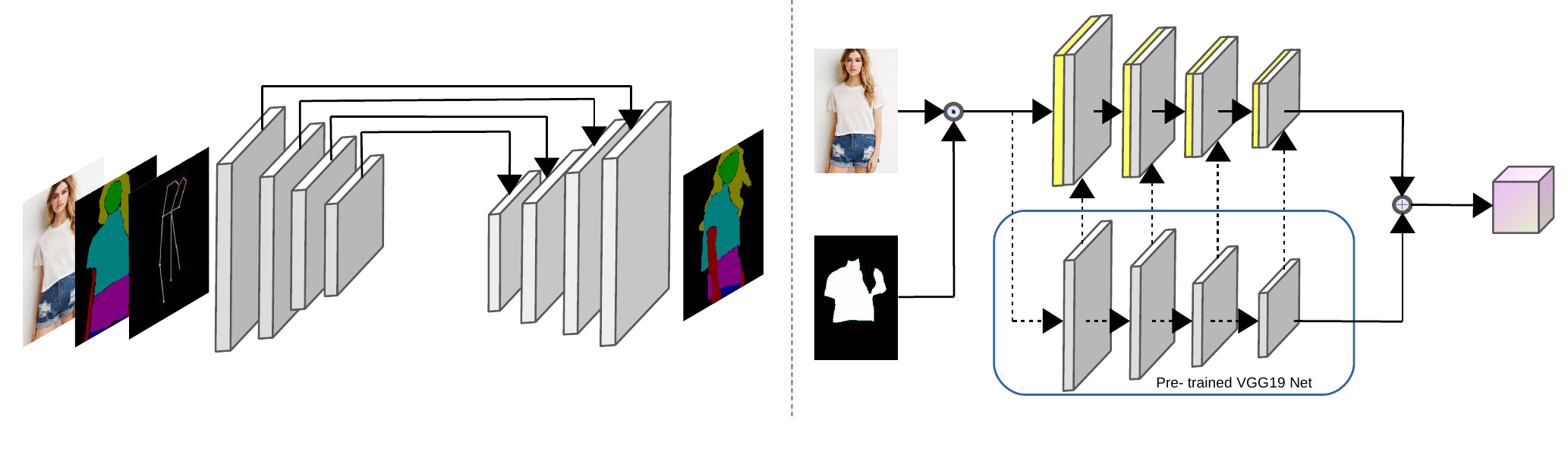}%设置图片的大小和图片的位置
	\caption{Left :pose-guided semantic map generation. Right :Details of decomposed attribute encoder in our generator. }%设置图片的名称
	\label{pose-guided semantic parsing generation fig}%设置图片的简称
\end{figure}

In this module, we try to use the source image $I_s$, the source semantic map $M_s$ and the keypoint-based source pose $P_s$ to generate the target semantic map. As shown in Figure \ref{pose-guided semantic parsing generation fig} (left), $I_s$, $M_s$, $P_s$ is used as the input of network $G_{parsing}$, which uses a Unet-based network structure, and the output is the predicted target semantic map $\hat{S}_t=G_{parsing}(I_s,M_s,P_s)$ by minimizing the pixel-wise $L1$ loss between $S_t$ and $\hat{S}_t$. Note that if we directly use keypoint-based pose , we cannot accurately establish the correspondence between the target pose and the source image. We address this problem in a coarse-to-fine way by predicting the target semantic map. Predicting target semantic map can not only provide effective structural constraints in the generation process, but also help establish the accurate dense correspondence between the  target pose and the source image.

%\begin{figure}[!ht]
%\centering%设置内容居中
%\includegraphics[scale=1]{figures/目标语义图生成.pdf}%设置图片的大小和图片的位置
%\includegraphics[scale=1.0]{figures/golfer.eps}
%\caption{pose-guided semantic map generation. }%设置图片的名称
%\label{pose-guided semantic parsing generation fig}%设置图片的简称
%\end{figure} 

\subsection{Generator}
Figure \ref{generatorfig} shows the architecture of the generator. During traing, the inputs of genetator are the target pose  $P_t$, the target semantic map $S_t$, the source semantic map $M_s$ and the source image $I_s$, and the output is the synthesized image $I_g=G(P_t,S_t,M_s,I_s)$ with the texture of $I_s$ and the pose of $I_t$. At test time, we replace $S_t$ with $\hat{S}_t$ as input of network since $S_t$ is not available. In fact, we can treat the target image as a deformed version of the source image, which means that the pixels on the target image can find the correspondening pixels on the source image. In order to find the dense correspondence, our generator encode the target pose and the source image into two latent codes by two encoders,called Pose Encoder (Section\ref{Pose Encoder}) and  Decomposed Attribute Encoder (Section\ref{Decomposed Attribute Encoder}). Then, dense correspondence between the target pose and the source image can be established within shared domain (Section\ref{Correspondence within shared domain}). Finally, texture renderer (Section\ref{Texture Renderer}) take warped source image feature and  target pose feature as inputs to generate final result.

\subsubsection{Pose encoder}\label{Pose Encoder}
Usually, the dense correspondence is established in the feature domain through a pre-trained model. According to this idea, we first map the target pose and the source image to the shared domain. Different from the previous method, which is to establish the correspondence between two images in the same domain, we try to establish the correspondence between pose representation and image in different domains. Specifically, Pose Encoder is composed of four down-sampling layers, which map the pose representation into the feature code $f_p$. the pose representation consist of the keypoint-based target pose $P_t$ and the target semantic map $S_t$. The reason why we use the target semantic map as a pose representation is that we think that the keypoint-based target pose is too sparse to accurately represent the correct pose, therefore, we use semantic map with richer information as pose representation to improve the accuracy of pose representation, which can also provide structural constraints during the generation process and help to establish the accurate dense correspondence between the  target pose and the source image.

\subsubsection{Decomposed attribute encoder}\label{Decomposed Attribute Encoder}
In order to more accurately extract the feature of source image and achieve the task of clothing-guided person image generation, we use decomposed attribute encoder for this part, inspired by \cite{men2020controllable}.  Figure \ref{pose-guided semantic parsing generation fig} (right) shows the architecture of decomposed attribute encoder. As mentioned before, each channel of $M_s$ corresponds to an attribute of the human body, so we first uses heatmap $M_s$ to extract correspondence attributes $I_s^i$ of $I_s$ by  
\begin{equation}
I_s^i=I_s \odot M_s^i
\end{equation} 
where $\odot$ denotes element-wise product, $M_s^i$ denotes the channel $i$ of $M_s$. afterthat, $I_s^i$ is mapped to texture code by pretarined VGG19-guided texture encoder $T_{enc}$. 
\begin{equation}
	f_s^i=T_{enc}(I_s^i)
\end{equation}
Finally, all texture code $f_s^i,i=1...K$ are concatenated in channel-wise to get full texture code $f_s$. Different from \cite{men2020controllable}, however,which use a avgpooling layer after each $f_s^i$, We modify the encoder network structure and cancel pooling, because we think pooling will cause a lot of information to be lost. What's more, in order to extract the feature of the texture code, \cite{men2020controllable} uses a fully connected layer to extract the affine parameters required by the AdaIN layer, which will greatly increase the parameters of the network and reduce the efficiency of the network. On the contrary, we use a convolutional network as the fushion module, which makes the model more lightweight.

\begin{figure}[!ht]
	\centering%设置内容居中
	\includegraphics[scale=0.75]{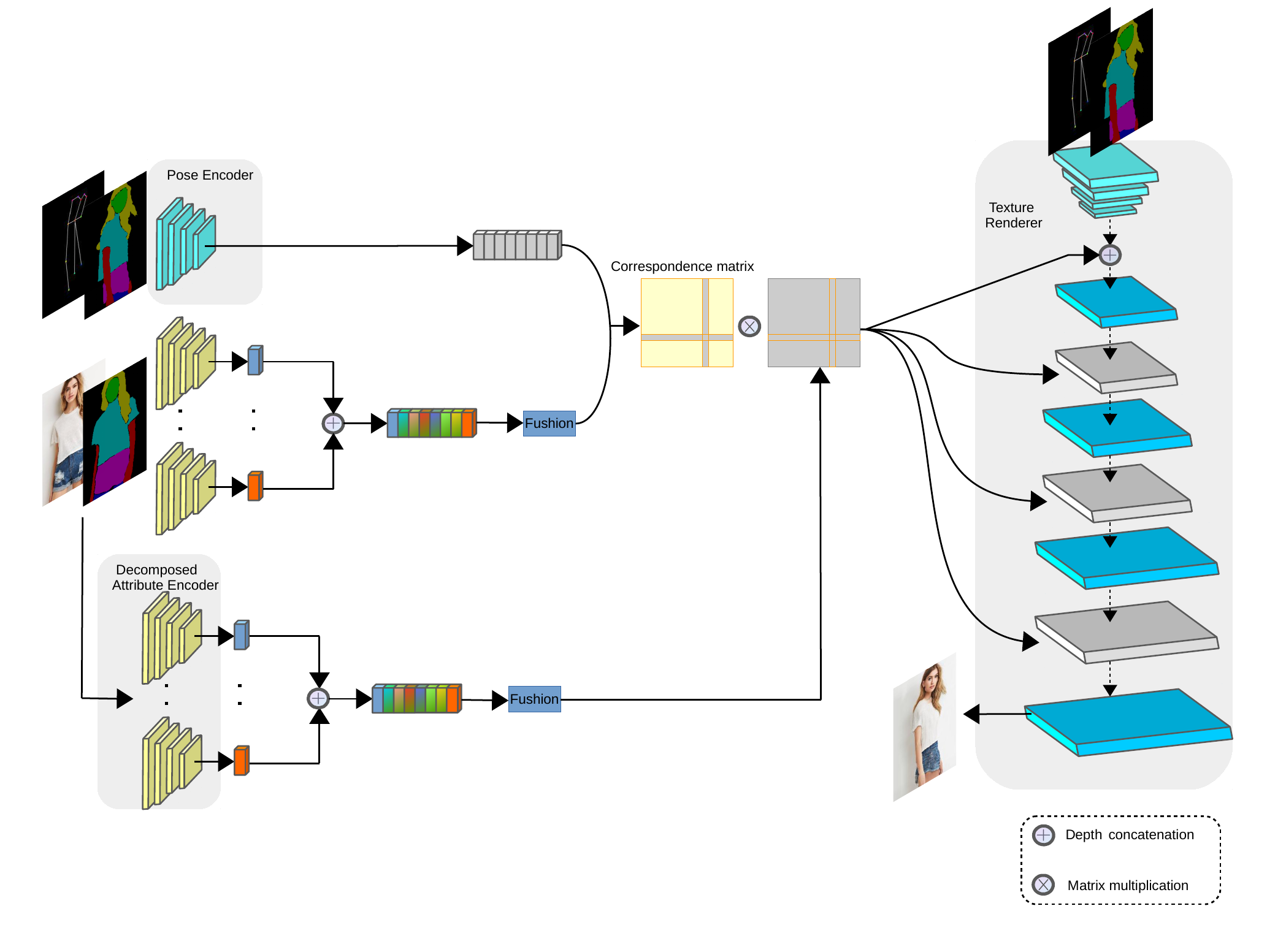}%设置图片的大小和图片的位置
	\caption{ An overview of the network architecture of our generator. During training, given inputs: $P_t$, $\hat{S}_t$, $M_s$ and $I_s$, we first use pose encoder and decomposed attribute encoder to map the pose and the source image to shared domain. Then, whthin the shared domain, we can establish dense correspondence between the target pose and the source image. The source image is embedded into the latent space via another independent decomposed attribute encoder, and warped source image feature can be obtained according to the dense correspondence. Finally, texture renderer generates output image based warped source image feature and target pose feature.}%设置图片的名称
	\label{generatorfig}%设置图片的简称
\end{figure} 
%\begin{figure}[!ht]
%	\centering%设置内容居中
%	\includegraphics[scale=0.5]{figures/属性提取模块.pdf}%设置图片的大小和图片的位置
	%\includegraphics[scale=1.0]{figures/golfer.eps}
%	\caption{ Details of decomposed attribute encoder in our generator.}%设置图片的名称
%	\label{Decomposed Attribute Encoderfig}%设置图片的简称
%\end{figure}

\subsubsection{Dense correspondence within shared domain}\label{Correspondence within shared domain}
After mapping the target pose and the source image into the shared domain, we get pose feature $f_p \in \mathbb{R}^{c \times h \times w}$ and texture code $f_s \in \mathbb{R}^{c \times h \times w}$, where $h,w$ are feature spatial size and $c$ is the channel wise demension. We then reshape $f_p \in \mathbb{R}^{c \times (hw)}$ and $f_s \in \mathbb{R}^{c \times (hw)}$. After that we can establish the correspondence between the target pose and the source image feature by 
\begin{equation}
	C(i,j)=\frac{(f_p(i)-\mu(f_p))^T\cdot(f_s(i)-\mu(f_s))}
	{|| f_p(i)-\mu(f_p) ||_2\cdot|| f_s(i)-\mu(f_s) ||_2}
\end{equation}
$C\in \mathbb{R}^{hw\times hw}$ is called correspondence matrix, whose elements $C(i,j)$ meature the similarity of $f_p$ at point $i$ and $f_s$ at point $j$. $\mu(f_p)$ and $\mu(f_s)$ represent mean value. Then, we can use the correspondence matrix to warp the source image feature by
\begin{equation}
f_{s\rightarrow t}(i)=\sum_j \mathop{softmax}_j (C(i,j))\cdot \bar{f}_s(j)
\end{equation}  
We use an independent decomposed attribute encoder to map the source image to shared domain and then apply reshape operation to get $\bar{f}_s(j) \in \mathbb{R}^{(hw) \times c}$.
Here we refer to the method of \cite{zhang2020cross} to establish the correspondence between the target pose and the source image. But unlike \cite{zhang2020cross}, which establishes the one-to-one correspondence between the semantic map and the exemplar image by selecting the largest value in each row of correspondence matrix. We think that duo to the  largely deformation of pose, there may be a one-to-many correspondence between the target pose and the source image, that is to say, there is a correspondence between a point on the target image and a point on the source image and its neighborhood. It's benefit if we can also establish this relationship, so we save the complete correspondence matrix. In addition, the other difference is that \cite{zhang2020cross} uses bilinear interpolation to downsample the source image to a specific size, whereas we use an independent decomposed attribute encoder to map the source image to specific size. In fact, the establishment of correspondence and the use of decomposed attribute encoder play a mutually promoting role: the establishment of correspondence can solve the problem of misalignment between the source image and the target image, so that the network can generate more realistic images, on the other hand, the use of decomposed attribute encoder can better extract image features to establish a more accurate dense correspondence.

\subsubsection{Texture Renderer}\label{Texture Renderer}
After obtaining the warped source image feature, we need to use this information to generate the final output. Figure \ref{generatorfig} shows the architecture of texture renderer. In order to better preserve the information of texture feature, we have borrowed from the SPADE network structure \cite{park2019semantic} . The texture features of different sizes are used as the input of the SPADE module. But we need to emphasize that: (1) As opposed to \cite{park2019semantic}, we replaced the BN layer in the SPADE module with the IN layer, because in the image generation task, different images have different styles, therefore, the IN layer that performs feature statistics on a single channel of a single instance is more suitable. (2) We did not use  constant code as the input of texture renderer, but target pose features. Compared to the constant code, the target pose feature has richer information, which can speed up the convergence of the model. What's more, target pose features can provide position constraints for texture features, making the dense corresponding more accurate, leading to more realistic generated image.

\subsection{Discriminator}\label{Discriminator}
Inspired by \cite{wang2018high}, We combine GAN loss and feature loss function in the discriminator to achieve a stable training effect. The input of the discriminator is the real image and the generated image. We calculate the feature loss function at the output of each layer of the discriminator, and calculate the GAN loss function in the last layer.

%Figure \ref{Discriminator} shows the architecture of Discriminator, 
%\begin{figure}[!ht]
%	\centering%设置内容居中
%	\includegraphics[scale=0.7]{figures/判别器.pdf}%设置图片的大小和图片的位置
	%\includegraphics[scale=1.0]{figures/golfer.eps}
%	\caption{ Details of discriminator.}%设置图片的名称
%	\label{Discriminator}%设置图片的简称
%\end{figure}

\subsection{Objective Functions}\label{Objective Functions}
 
Due to the complexity of the task, we apply a joint loss to train our network, including adversarial loss, feature loss, reconstruction loss, perceptual loss, Contextual loss and correspondence loss,  written as follows. 
\begin{equation}
	\mathcal{L}_{total}=\lambda_{adv}\mathcal{L}_{adv}+\lambda_{fea}\mathcal{L}_{fea}+\lambda_{rec}\mathcal{L}_{rec}+\lambda_{per}\mathcal{L}_{per}+\lambda_{con}\mathcal{L}_{con}+\lambda_{cor}\mathcal{L}_{cor}
\end{equation}
where $\lambda_{adv}$, $\lambda_{fea}$, $\lambda_{rec}$, $\lambda_{per}$, $\lambda_{con}$, $\lambda_{cor}$ denote the weights of corresponding losses, respectively. The goal of adversarial loss is to make the distribution of the generated image as close as possible to the distribution of the real image, which is defined as 
\begin{equation}
	\mathcal{L}_{adv}=\mathbb{E}[\log (1-D(G(I_s,S_s,S_t,P_t))]+\mathbb{E}[\log D(I_t)]
\end{equation}
The feature loss can be written as 
\begin{equation}
	\mathcal{L}_{fea}=\sum_{i=0}^{n}\alpha_i|| D_i(I_g)-D_i(I_t) ||_1
\end{equation} 
where $D_i$ denotes the $i-$th ($i=0,1,2$) layer feature from discriminator, $\alpha_i$ denotes the weight of the feature loss of each layer.
The reconstruction loss is used to penalize the difference between the generated image and the real image at the pixel level:
 \begin{equation}
 	\mathcal{L}_{rec}=||I_g-I_t||_1
 \end{equation}
 In addition, we also used the perceptual loss to match the deep features of the image, which is effective in image generation tasks.
 \begin{equation}
 	\mathcal{L}_{per}=|| \phi_l(I_g)-\phi_l(I_t) ||_1
 \end{equation}
where $\phi_l$ denotes the output of $l-$th layer from the pretarined VGG-19 model.
We also adopt contextual loss proposed in \cite{mechrez2018contextual}, which is designed for image generation that naturally handles tasks with non-aligned training data and is very suitable for our task. The contexttual loss is defined as follow.
\begin{equation}
	\mathcal{L}_{con}=-\log (\mathop{CX}(\phi_l(x),\phi_l(y))
\end{equation}
where $CX$ denotes the contexttual similarity between feature. The detailed definition can be found in \cite{mechrez2018contextual}.
Finally, in order to ensure the correctness of the correspondence matrix, we use correspondence loss:
\begin{equation}
	\mathcal{L}_{cor}=|| f_{s \rightarrow t}-\phi_l(I_t)||_1
\end{equation}

\section{Experiments}
\subsection{Implementation Details}

\paragraph{Datasets.}We conduct experiments on the In-shop Cloths Retrieval Benchmask of the Deepfashion dataset \cite{liu2016deepfashion}, which contains 52,712 images of people with varying poses and appearances. We adopt the data division in \cite{men2020controllable}, using 10,1966 image pairs as training dataset and 8750 image pairs as test dataset to ensure that the same person does not appear in the training dataset and the test dataset, and crop the image with a resolution of $256 \times 256$ to remove the excess background, leaving a $256 \times 176$ area in the center.
\paragraph{Metrics.}We use Inception Score(IS) \cite{salimans2016improved}, Structural Similarity(SSIM) \cite{wang2004image}, Frechet Inception Distance(FID) \cite{heusel2017gans} and
Learned Perceptual Image Patch Similarity(LPIPS) \cite{zhang2018unreasonable} 
to quantitatively evaluate the quality of the generated image, which are commonly used evaluation metrics in image generation task. For IS and SSIM, a higher score is better. For FID and LPIPS, a lower score is better.
\paragraph{Network Implementation and Training Details.}
We implement our network based on pytorch, using four 2080Ti GPUs. Pose Encoder contains 4 down-sampling layers, decomposed attribute encoder contains 3 down-sampling layers, and texture renderer is composed of 3 SPADE ResBlks.  Discriminator is composed of three down-sampling layers. In each layer of the network, spectrum normalization \cite{miyato2018spectral} is used to stabilize network training. We use the Adams optimizer \cite{DBLP:journals/corr/KingmaB14} with $\beta_1=0.5$ and $\beta_2=0.999$. Inspired by TTUR \cite{heusel2017gans}, we set the initial learning rates of the generator and discriminator to 0.0002 and 0.0003, respectively. The batchsize is set to 4, the training process lasts for 30 epochs, and the learning rate is  linearly decayed to 0 after 15 epochs.

\begin{figure}[!ht]
	\centering%设置内容居中
	\includegraphics[scale=0.7]{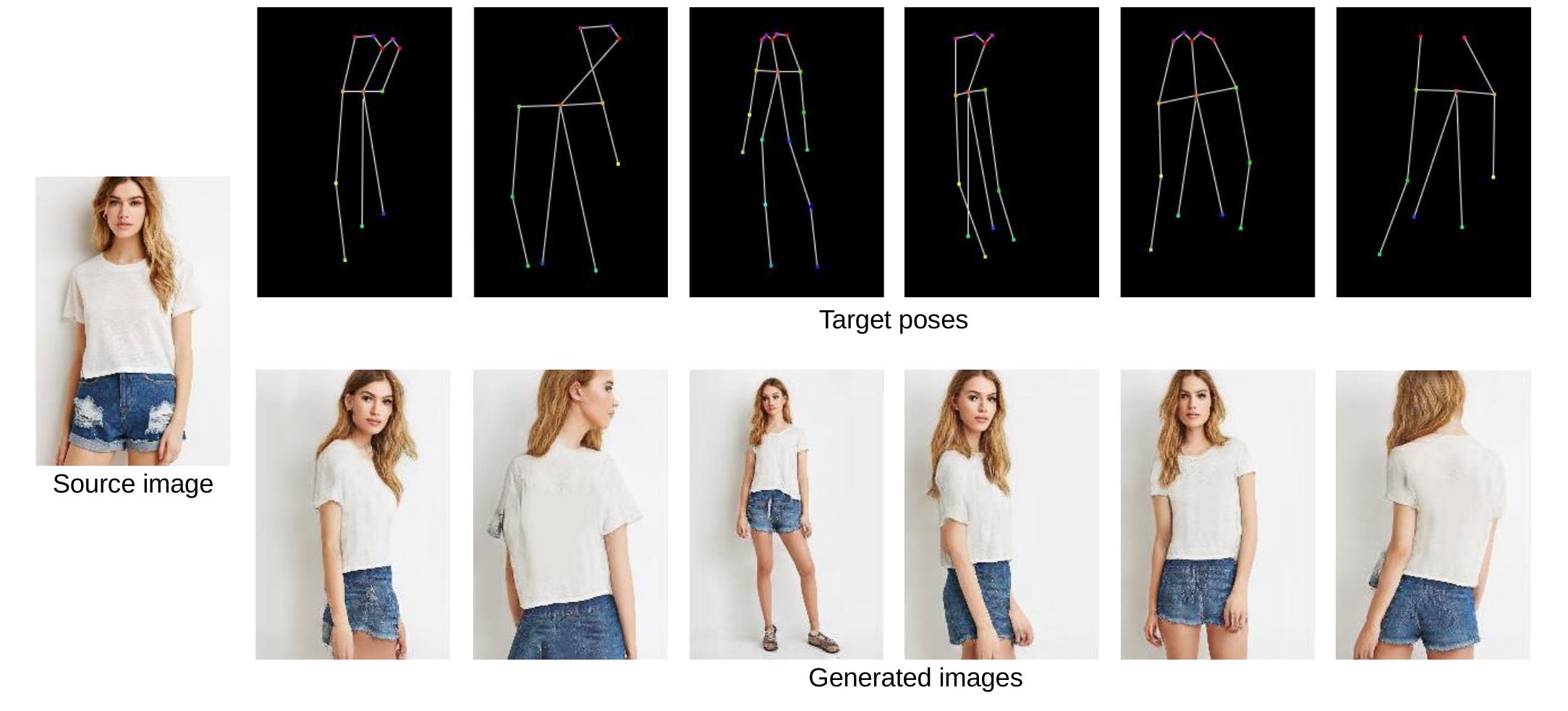}%设置图片的大小和图片的位置
	\caption{Results of synthesizing person images  in arbitrary poses at Deepfashion dataset}%设置图片的名称
	\label{Results of synthesizing person images}%设置图片的简称
\end{figure}

\subsection{Pose-guided image synthesis results}
The result of pose-guided image synthesis is shown in Figure \ref{Results of synthesizing person images}. Given an source image and any target pose, our method can convert the pose to the target pose while maintaining the texture of the source image.

\begin{figure}[!ht]
	\centering%设置内容居中
	\includegraphics[scale=0.7]{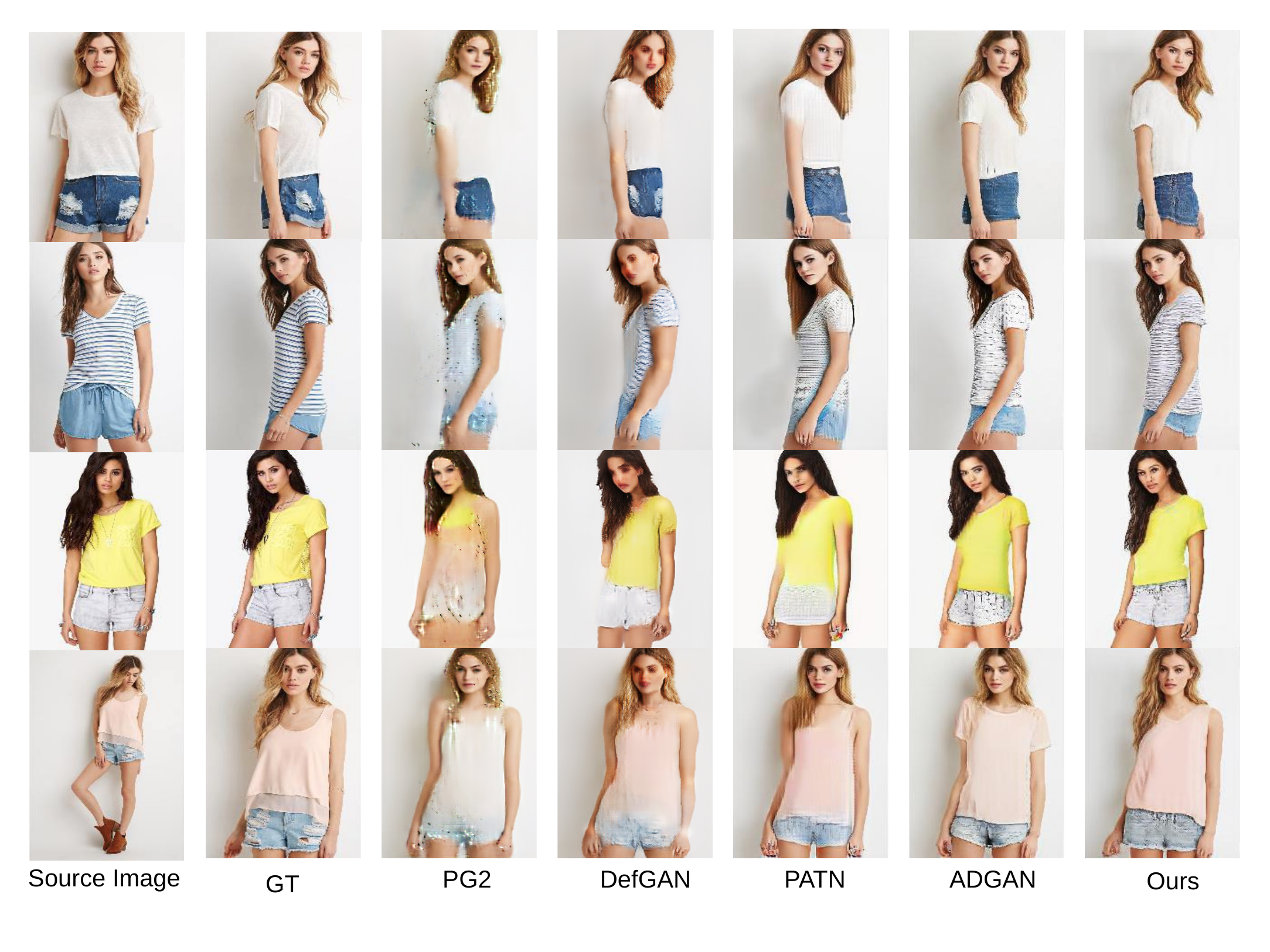}%设置图片的大小和图片的位置
	\caption{Qualitative comparison with state-of-the art methods.}%设置图片的名称
	\label{Qualitative comparison against state-of-the art methods}%设置图片的简称
\end{figure}

\subsubsection{Qualitative result} In addition, we select some  state-of-the-art pose
transfer methods as baselines to compare with our method , including PG2 \cite{ma2017pose}, DefGAN \cite{siarohin2018deformable}, PATN \cite{zhu2019progressive} and ADGAN  \cite{men2020controllable}. Among them, Def-GAN and PATN model the deformation between the source image and the target image, but PG2 and ADGAN do not consider it. The result of the comparison method is provided by authors or obtained by open source code and pre-trained model. The qualitative comparison result is shown in Figure \ref{Qualitative comparison against state-of-the art methods}. We can learn from the results, on the one hand, compared with PG2 and ADGAN, which generate the a blurry and coarse results without considering the establishment of the dense correspondence between the source image and the target image, while our method has the correct pose and more detailed texture results due to model the deformation. On the other hand, compared with other methods that model deformation, such as DefGAN and PATN, our method can also generate more natural and realistic results due to the mutual promotion of dense correspondence and decomposed attribute modules, especially on faces identity and hair texture.

%that the existing method,such as PG2 and ADGAN, generate a blurry and coarse image without considering the establishment of the dense corresponding between the source image and the target image. In contrast, the image generated by our method can not only ensure the correct pose, but also has more detail texture, especially the facial identity and hair attributes

\begin{table}
	\caption{Quantitative comparison with state-of-the art methods.
		Deform indicates if the method models deformation} 
	\centering
	\begin{tabular}{llllll}
		\toprule
		%	\multicolumn{3}{c}{Part}                   \\
		\cmidrule(r){1-6}
		Model   &Deform  & SSIM $\uparrow$     & IS $\uparrow$ &FID $\downarrow$  & LPIPS $\downarrow$ \\
		\midrule
		PG2 \cite{ma2017pose}(NIPS2017)
		& \XSolidBrush &  0.773  & 3.202 &47.713&0.245   \\
		Def-GAN \cite{siarohin2018deformable}(CVPR2018) 
		&\Checkmark & 0.756  & 3.439  &26.430&0.209    \\
		PATN \cite{zhu2019progressive}(CVPR2019)     
		&\Checkmark & 0.771  & 3.203 &19.822&0.196 \\
		ADGAN \cite{men2020controllable}(CVPR2020)     
		&\XSolidBrush & 0.770  & 3.392  &13.009&0.177 \\
		\midrule
		
		%w/o parsing数据需要更新
		w/o parsing    &\Checkmark & 0.727       & 3.461  &45.404&0.242 \\
		
		w/o pose feature   &\Checkmark & 0.806       & 3.410  &13.487 &0.134 \\
		
		%w/o cprrespondence loss数据需要更新
		w/o GAN loss &\Checkmark & 0.808       & 3.448  & 13.439 &0.132\\
		
		%w/o correspondence matrix &\XSolidBrush & 0.812      & 3.442  &11.791&0.131\\
		
		Ours(full)   & \Checkmark & \textbf{0.814}   & \textbf{3.538} &\textbf{11.385}&\textbf{0.132} \\
		
		\midrule
		
		Real Data & - &1.000  &3.898	  &0.000 &0.000 \\
		\bottomrule
	\end{tabular}
	\label{Quantitative  Results}
\end{table}

\subsubsection{Quantitative result}

In order to verify the effectiveness of our method more objectively, we conducted quantitative experiments to compare our method with state-of-the art methods. The results of the quantitative comparison are shown in Table \ref{Quantitative  Results}. It can be seen from the table that our method is better than state-of-the art methods in all three evaluation Metrics, increasing the best IS score from 0.771 to 0.814, improving the best SSIM score from 3.439 to 3.538 and reducing the best FID score from 13.009 to 11.385.
The results of qualitative experiments further verify the effectiveness of our method.

\subsubsection{Ablation study}
In order to verify the influence of the important part of the proposed method on the final result, we conducted ablation study. Our ablation study is divided into the following parts: \textbf{w/o parsing:} In order to verify that semantic map can provide a richer pose information, we did not use semantic map as a target pose representation, but only a keypoint-based pose representation. \textbf{w/o pose feature:} We replace the pose feature input of texture renderer with the constant input to illustrate the role of pose feature in the decoding process. \textbf{w/o GAN loss:} We retrain our model without using GAN loss to illustrate the impact of GAN loss. %

\begin{figure}[!ht]
	\centering%设置内容居中
	\includegraphics[scale=0.5]{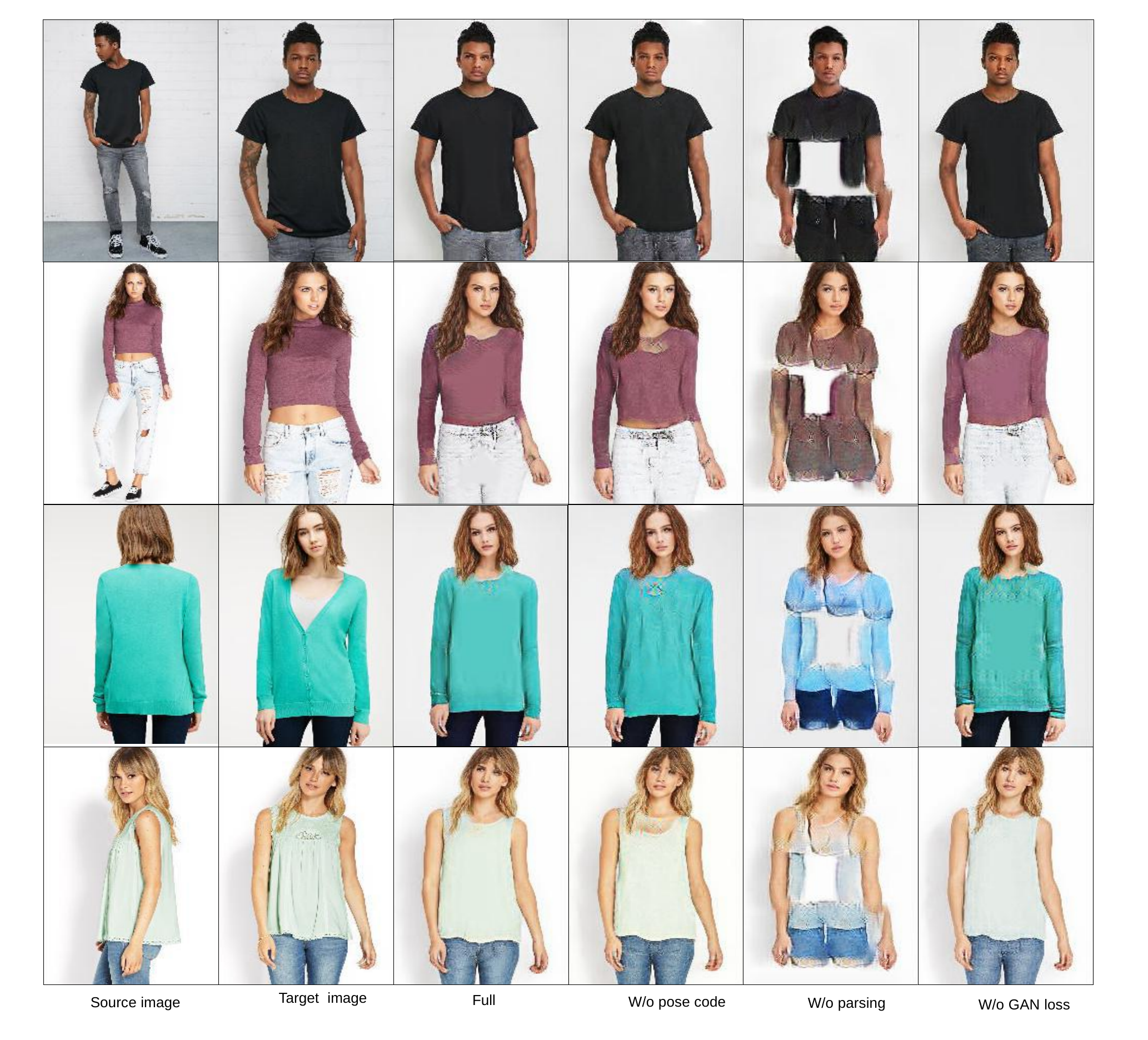}%设置图片的大小和图片的位置
	\caption{Results of ablation studies.}%设置图片的名称
	\label{Ablation studies}%设置图片的简称
\end{figure}

%再做一个消融实验：将correspondence matrix去掉直接将目标姿态编码和原图像编码作为解码器的输入，看一下实验的效果怎么样，因此说明correspondence matrix的有效性

The results of the ablation study are shown in Table \ref{Quantitative  Results} and Figure \ref{Ablation studies}. we can learn from the results, 
We first compare the results with using semantic map and those without it. We can learn that the target semantic map generation can provide effective structural constraints during the image generation process and improve the quality of the generated images. After that, we can know from the comparison result of the result of using the constant input and the result of using the target pose feature as the input that the target pose feature as input can effectively guide the rendering of texture features. Moreover, under the constraint of GAN loss, the network can generate more realistic image. We can learn from the results: the semantic map has a major impact on the final result, and not using pose code and GAN loss will slightly reduce the quality of the generated image.

\subsection{Clothing-guided image synthesis results}
\begin{figure}[!ht]
	\centering%设置内容居中
	\includegraphics[scale=0.7]{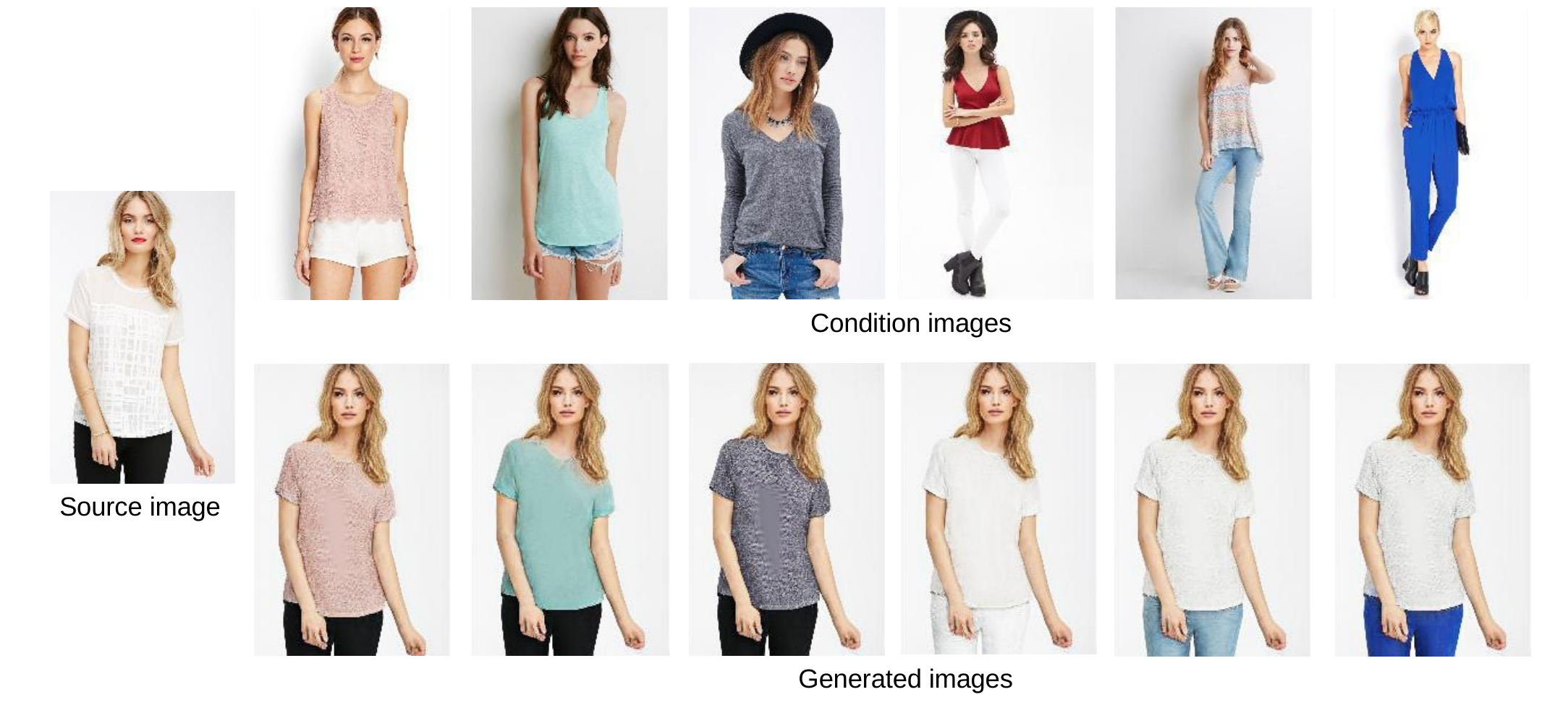}%设置图片的大小和图片的位置
	\caption{Results of synthesizing person images with controllable component attributes.}%设置图片的名称
	\label{Results of synthesizing person images with controllable component attributes}%设置图片的简称
\end{figure}

As described in Section \ref{Decomposed Attribute Encoder}, decomposed attribute encoder encodes different attributes through a shared encoding module, and finally combines all the codes to generate images. Through decomposed attribute encoder, we can extract the desired clothing attributes from different source images and combine them to achieve clothing-guided image synthesis. As shown in Figure \ref{Results of synthesizing person images with controllable component attributes}, given an source image, the first row represents the conditional image with the desired clothing attributes, and the second row represents the generated image. In the first three columns, we changed upper clothes of the source image according to the desired clothing attributes. In the last three columns, we changed  pants of the source image according to the desired clothing attributes.

%可以再补充和ADGAN的clothing-guided 的比较。

\section{Conclusion}
In this paper, we propose a new network architecture for controllable person image generation, which can be applied to pose-guided person image generation and clothing-guided person image generation. Our method generates target semantic map under the guidance of target pose, which can provide effective structural constraints during the generation process. In order to effectively combine the pose-guided person image generation task and the clothing-guided person image generation task, we use two independent encoders to map the target pose and the source image to the shared domain. In the shared domain, we can establish the dense corresponding between the target pose and the source image. Under the guidance of dense corresponding , we can generate  realistic and natural results. The experimental results show the effectiveness of our method in two tasks.

\bibliographystyle{unsrt}
%\newpage

\bibliography{references}  %%% Remove comment to use the external .bib file (using bibtex).

\begin{thebibliography}{10}

\bibitem{ma2017pose}
Liqian Ma, Xu~Jia, Qianru Sun, Bernt Schiele, Tinne Tuytelaars, and Luc
  Van~Gool.
\newblock Pose guided person image generation.
\newblock In {\em Advances in neural information processing systems}, pages
  406--416, 2017.

\bibitem{men2020controllable}
Yifang Men, Yiming Mao, Yuning Jiang, Wei-Ying Ma, and Zhouhui Lian.
\newblock Controllable person image synthesis with attribute-decomposed gan.
\newblock In {\em Proceedings of the IEEE/CVF Conference on Computer Vision and
  Pattern Recognition}, pages 5084--5093, 2020.

\bibitem{ma2018disentangled}
Liqian Ma, Qianru Sun, Stamatios Georgoulis, Luc Van~Gool, Bernt Schiele, and
  Mario Fritz.
\newblock Disentangled person image generation.
\newblock In {\em Proceedings of the IEEE Conference on Computer Vision and
  Pattern Recognition}, pages 99--108, 2018.

\bibitem{esser2018variational}
Patrick Esser, Ekaterina Sutter, and Bj{\"o}rn Ommer.
\newblock A variational u-net for conditional appearance and shape generation.
\newblock In {\em Proceedings of the IEEE Conference on Computer Vision and
  Pattern Recognition}, pages 8857--8866, 2018.

\bibitem{siarohin2018deformable}
Aliaksandr Siarohin, Enver Sangineto, St{\'e}phane Lathuiliere, and Nicu Sebe.
\newblock Deformable gans for pose-based human image generation.
\newblock In {\em Proceedings of the IEEE Conference on Computer Vision and
  Pattern Recognition}, pages 3408--3416, 2018.

\bibitem{ren2020deep}
Yurui Ren, Xiaoming Yu, Junming Chen, Thomas~H Li, and Ge~Li.
\newblock Deep image spatial transformation for person image generation.
\newblock In {\em Proceedings of the IEEE/CVF Conference on Computer Vision and
  Pattern Recognition}, pages 7690--7699, 2020.

\bibitem{li2019dense}
Yining Li, Chen Huang, and Chen~Change Loy.
\newblock Dense intrinsic appearance flow for human pose transfer.
\newblock In {\em Proceedings of the IEEE Conference on Computer Vision and
  Pattern Recognition}, pages 3693--3702, 2019.

\bibitem{zhu2019progressive}
Zhen Zhu, Tengteng Huang, Baoguang Shi, Miao Yu, Bofei Wang, and Xiang Bai.
\newblock Progressive pose attention transfer for person image generation.
\newblock In {\em Proceedings of the IEEE Conference on Computer Vision and
  Pattern Recognition}, pages 2347--2356, 2019.

\bibitem{goodfellow2014generative}
Ian Goodfellow, Jean Pouget-Abadie, Mehdi Mirza, Bing Xu, David Warde-Farley,
  Sherjil Ozair, Aaron Courville, and Yoshua Bengio.
\newblock Generative adversarial nets.
\newblock In {\em Advances in neural information processing systems}, pages
  2672--2680, 2014.

\bibitem{kingma2013auto}
Diederik~P Kingma and Max Welling.
\newblock Auto-encoding variational bayes.
\newblock In {\em International Conference on Learning Representations}, 2014.

\bibitem{Radford2016UnsupervisedRL}
Alec Radford, Luke Metz, and Soumith Chintala.
\newblock Unsupervised representation learning with deep convolutional
  generative adversarial networks.
\newblock {\em CoRR}, abs/1511.06434, 2016.

\bibitem{mirza2014conditional}
Mehdi Mirza and Simon Osindero.
\newblock Conditional generative adversarial nets.
\newblock {\em arXiv preprint arXiv:1411.1784}, 2014.

\bibitem{wang2018high}
Ting-Chun Wang, Ming-Yu Liu, Jun-Yan Zhu, Andrew Tao, Jan Kautz, and Bryan
  Catanzaro.
\newblock High-resolution image synthesis and semantic manipulation with
  conditional gans.
\newblock In {\em Proceedings of the IEEE conference on computer vision and
  pattern recognition}, pages 8798--8807, 2018.

\bibitem{park2019semantic}
Taesung Park, Ming-Yu Liu, Ting-Chun Wang, and Jun-Yan Zhu.
\newblock Semantic image synthesis with spatially-adaptive normalization.
\newblock In {\em Proceedings of the IEEE Conference on Computer Vision and
  Pattern Recognition}, pages 2337--2346, 2019.

\bibitem{karras2019style}
Tero Karras, Samuli Laine, and Timo Aila.
\newblock A style-based generator architecture for generative adversarial
  networks.
\newblock In {\em Proceedings of the IEEE conference on computer vision and
  pattern recognition}, pages 4401--4410, 2019.

\bibitem{karras2020analyzing}
Tero Karras, Samuli Laine, Miika Aittala, Janne Hellsten, Jaakko Lehtinen, and
  Timo Aila.
\newblock Analyzing and improving the image quality of stylegan.
\newblock In {\em Proceedings of the IEEE/CVF Conference on Computer Vision and
  Pattern Recognition}, pages 8110--8119, 2020.

\bibitem{zhou2019text}
Xingran Zhou, Siyu Huang, Bin Li, Yingming Li, Jiachen Li, and Zhongfei Zhang.
\newblock Text guided person image synthesis.
\newblock In {\em Proceedings of the IEEE Conference on Computer Vision and
  Pattern Recognition}, pages 3663--3672, 2019.

\bibitem{cao2017realtime}
Zhe Cao, Tomas Simon, Shih-En Wei, and Yaser Sheikh.
\newblock Realtime multi-person 2d pose estimation using part affinity fields.
\newblock In {\em Proceedings of the IEEE conference on computer vision and
  pattern recognition}, pages 7291--7299, 2017.

\bibitem{gong2017look}
Ke~Gong, Xiaodan Liang, Dongyu Zhang, Xiaohui Shen, and Liang Lin.
\newblock Look into person: Self-supervised structure-sensitive learning and a
  new benchmark for human parsing.
\newblock In {\em Proceedings of the IEEE Conference on Computer Vision and
  Pattern Recognition}, pages 932--940, 2017.

\bibitem{zhang2020cross}
Pan Zhang, Bo~Zhang, Dong Chen, Lu~Yuan, and Fang Wen.
\newblock Cross-domain correspondence learning for exemplar-based image
  translation.
\newblock In {\em Proceedings of the IEEE/CVF Conference on Computer Vision and
  Pattern Recognition}, pages 5143--5153, 2020.

\bibitem{mechrez2018contextual}
Roey Mechrez, Itamar Talmi, and Lihi Zelnik-Manor.
\newblock The contextual loss for image transformation with non-aligned data.
\newblock In {\em Proceedings of the European Conference on Computer Vision
  (ECCV)}, pages 768--783, 2018.

\bibitem{liu2016deepfashion}
Ziwei Liu, Ping Luo, Shi Qiu, Xiaogang Wang, and Xiaoou Tang.
\newblock Deepfashion: Powering robust clothes recognition and retrieval with
  rich annotations.
\newblock In {\em Proceedings of the IEEE conference on computer vision and
  pattern recognition}, pages 1096--1104, 2016.

\bibitem{salimans2016improved}
Tim Salimans, Ian Goodfellow, Wojciech Zaremba, Vicki Cheung, Alec Radford, and
  Xi~Chen.
\newblock Improved techniques for training gans.
\newblock In {\em Advances in neural information processing systems}, pages
  2234--2242, 2016.

\bibitem{wang2004image}
Zhou Wang, Alan~C Bovik, Hamid~R Sheikh, and Eero~P Simoncelli.
\newblock Image quality assessment: from error visibility to structural
  similarity.
\newblock {\em IEEE transactions on image processing}, 13(4):600--612, 2004.

\bibitem{heusel2017gans}
Martin Heusel, Hubert Ramsauer, Thomas Unterthiner, Bernhard Nessler, and Sepp
  Hochreiter.
\newblock Gans trained by a two time-scale update rule converge to a local nash
  equilibrium.
\newblock In {\em Advances in neural information processing systems}, pages
  6626--6637, 2017.

\bibitem{zhang2018unreasonable}
Richard Zhang, Phillip Isola, Alexei~A Efros, Eli Shechtman, and Oliver Wang.
\newblock The unreasonable effectiveness of deep features as a perceptual
  metric.
\newblock In {\em Proceedings of the IEEE conference on computer vision and
  pattern recognition}, pages 586--595, 2018.

\bibitem{miyato2018spectral}
Takeru Miyato, Toshiki Kataoka, Masanori Koyama, and Yuichi Yoshida.
\newblock Spectral normalization for generative adversarial networks.
\newblock In {\em International Conference on Learning Representations}, 2018.

\bibitem{DBLP:journals/corr/KingmaB14}
Diederik~P. Kingma and Jimmy Ba.
\newblock Adam: {A} method for stochastic optimization.
\newblock In {\em International Conference on Learning Representations}, 2015.

\end{thebibliography}

\end{document}